\documentclass[preprint,5p,times,twocolumn]{elsarticle}

\usepackage{amssymb}
\usepackage{mdwmath}
\usepackage{mdwtab}
\usepackage{eqparbox}
\usepackage{url}
\usepackage{hyperref}
\usepackage{booktabs}
\usepackage{multirow}
\usepackage[cmex10]{amsmath}
\usepackage{color}
\usepackage{hhline}
\usepackage{comment}
\hypersetup{hidelinks}

\journal{Computer \& Graphics}

\begin{document}

\begin{frontmatter}



\title{Palette-based Color Transfer between Images}


\author[1]{Chenlei~Lv}

\author[2]{Dan Zhang\corref{cor1}}
\ead{danz@mail.bnu.edu.cn}

\address[1]{College of Computer Science and Software Engineering, Shenzhen University, China}
\address[2]{School of Computer Science, Qinghai Normal University, China}


\begin{abstract} 

As an important subtopic of image enhancement, color transfer aims to enhance the color scheme of a source image according to a reference one while preserving the semantic context. To implement color transfer, the palette-based color mapping framework was proposed. \textcolor{black}{It is a classical solution that does not depend on complex semantic analysis to generate a new color scheme. However, the framework usually requires manual settings, blackucing its practicality.} The quality of traditional palette generation depends on the degree of color separation. In this paper, we propose a new palette-based color transfer method that can automatically generate a new color scheme. With a redesigned palette-based clustering method, pixels can be classified into different segments according to color distribution with better applicability. {By combining deep learning-based image segmentation and a new color mapping strategy, color transfer can be implemented on foreground and background parts independently while maintaining semantic consistency.} The experimental results indicate that our method exhibits significant advantages over peer methods in terms of natural realism, color consistency, generality, and robustness.


\end{abstract}



\begin{keyword}



image enhancement, image recoloring, color transfer, color mapping, palette

\end{keyword}

\end{frontmatter}


\section{Introduction}

Following the development of digital media technology, researchers propose various enhanced frameworks to satisfy related requirements on digital display. Such frameworks include image denoising, super-resolution, low-light enhancement, color transfer, style transfer, etc. As an important subtopic, the color transfer framework attempts to learn the color scheme from the reference image to recolor the source one. It enhances the quality of color distribution without complex manual intervention. The framework is widely used in color management~\cite{usui2000neural}, medical image processing~\cite{he2022deconv}, camera systems~\cite{song2016sufficient}, and film post-production~\cite{huang2011temporal}.

{The pioneer work of color transfer was proposed by Reinhard in~\cite{reinhard2001color}.} It establishes the relationship between images in Lab color space, which provides a decoupled representation of color information. The main drawback of the work is that the semantic correspondences between images are ignored. Based on the pioneer work, many researchers propose improvement plans according to semantic analysis, such as patchmatch searching~\cite{hacohen2011non}, SIFT detection~\cite{park2016efficient}, CNN-based correspondence~\cite{he2018deep}, etc. Benefiting from the semantic analysis, such methods achieve more accurate color mapping while maintaining semantic consistency. {Compared to the pioneer work, these methods generate a more natural color scheme in the final result.} However, the methods depend on semantic correspondences between images. Once the images for color transfer lack correspondences, the performance of the methods will drop significantly.

To solve the problem, the palette-based color transfer methods~\cite{Greenfield2003image} are proposed. They implement the color mapping between images based on color segments represented by a palette. {Such a scheme has two significant advantages: independence from semantic correspondence and consistency in color distribution.} The palette-based color transfer learns the color segments from images to build the correspondence without complex semantic analysis. It avoids mapping errors produced by ambiguous semantic correspondences. The segment represents an accurate color-based continuous area, which provides consistency in color distribution and guarantees the natural property in the color transfer result. Based on these advantages, the palette-based color transfer is suitable for images without significant semantic correspondences. However, existing palette-based color transfer methods~\cite{zhang2021blind}\cite{zhang2017palette} require some manual inputs, which reduces practicality in related applications. In addition, the quality of palette generation depends on the color distribution of the image. Once the color distribution is relatively scattered for palette generation, the performance of the methods will necessarily degrade.

In this paper, we propose a new palette-based color transfer framework without any manual inputs. It is constructed by three components: palette-based clustering, color mapping strategy, and lighting optimization. {The palette-based clustering aims to establish different color segments for the image according to color distribution.} The proposed clustering method uses histogram analysis in the Lab color space to generate the palette. Based on the achieved palette, we propose the color mapping strategy to map palette-based color distribution from the reference image to the source one. The strategy maintains consistency in color distribution while aligning related colors based on the palette. To improve mapping accuracy with better correspondence, we utilize a deep learning-based segmentation method~\cite{chen2018encoder} to split foreground and background parts in images and implement color transfer to them independently. Then, we achieve a better color mapping result while maintaining color consistency. Finally, we optimize the illuminations of the source image according to the reference one as an optional action. This is used to avoid abnormal exposure. The proposed framework provides a reasonable solution for images without strict semantic correspondence requirements. It avoids discontinuity in color distribution and unnatural color mapping as much as possible. Compared to traditional palette-based color transfer or recoloring methods, our framework doesn't require manual input. To summarize, the contributions of this work include:

\begin{itemize}

\item We present a new palette-based clustering method to establish the palette for images. It improves the accuracy and applicability of palette generation.{Compared to traditional solutions like the k-means clustering-based method, our method doesn't require specifying clustering centers.} The palette can be generated automatically based on histogram analysis in the Lab color space

\item We propose a color mapping strategy to implement color transfer based on the achieved palette. The strategy aligns palette between images with precise chromatic aberration control. Combining a deep learning-based image segmentation, the color mapping can be implemented for foreground and background parts independently while keeping the color consistency.

\item We provide a lighting optimization as an optional action for reference images with abnormal exposure. The optimization can be regarded as a global adjustment according to illuminations. It enhances the naturalness of image after color transfer and avoids extreme exposure errors.

\end{itemize}

\begin{figure}
  \centering
  \includegraphics[width=\linewidth]{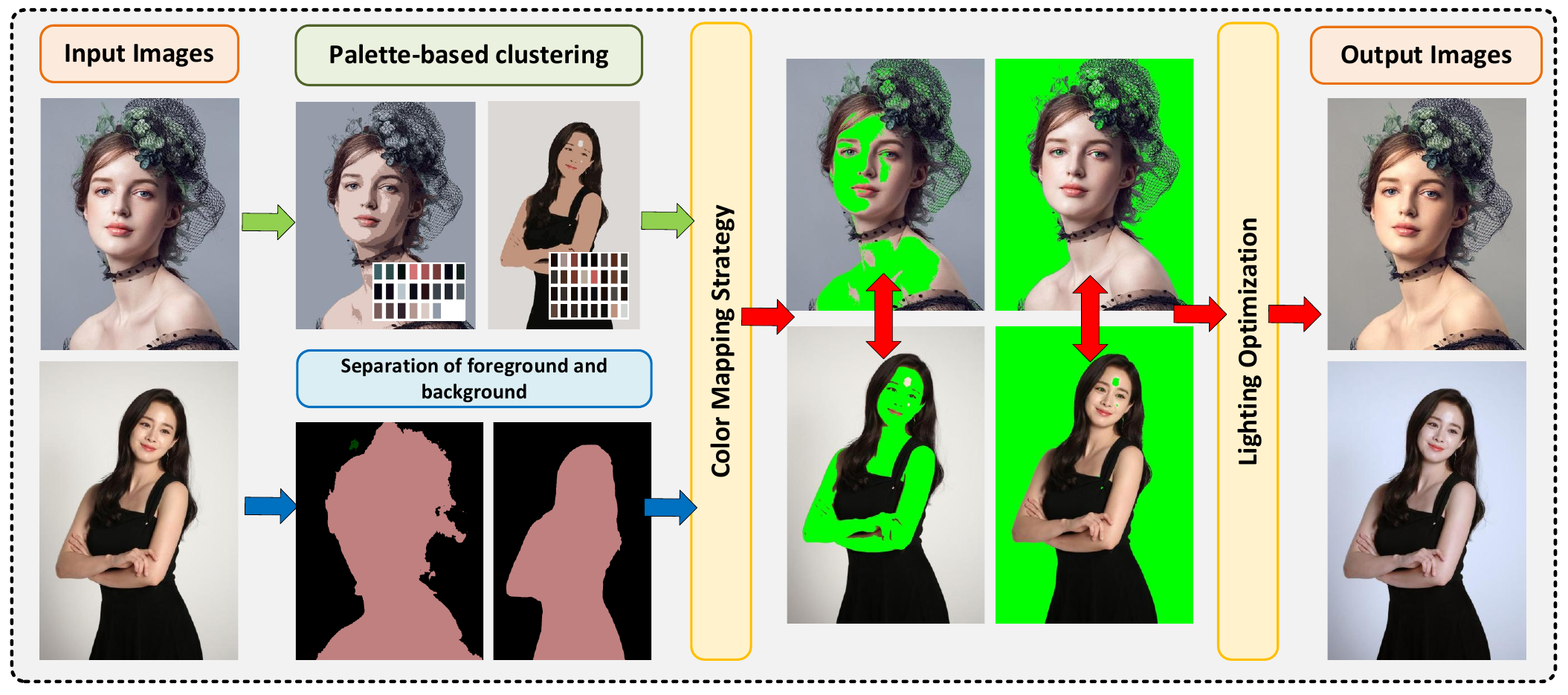}
  \caption{The pipeline of our framework.}
  \label{f1}
\end{figure}

The pipeline of our method is shown in Figure \ref{f1}. The rest of the paper is organized as follows. In Sec. 2, we review existing classical methods for color transfer. We introduce the palette-based clustering in Sec. 3, followed by the color mapping strategy and lighting optimization in Sec. 4 and Sec. 5. We demonstrate the effectiveness and efficiency of our method with extensive experimental evidence in Sec. 6, and Sec. 7 concludes the paper.

\section{Related Works}

According to different mechanisms of color transfer~\cite{lv2023color}, the related works can be classified into three categories: global color transfer, correspondence-based color transfer, and deep feature-based color transfer.

The global color transfer methods attempt to establish color mapping according to global color distribution. The pioneer work is proposed by Reinhard~\cite{reinhard2001color} in 2002. It constructs the color mapping in Lab color space which provides a decoupled representation of color information. Based on the same structure, Wang $et~al.$~\cite{wang2006effective} presented the color transfer scheme with B-spline control for video sequence. Kumar $et~al.$~\cite{kumar2008motion} designed a color transfer method in YCbCr color space for images with motion blur. Morovic $et~al.$~\cite{morovic2003accurate} uilized earth move distance (EMD) to implement histogram-based color transfer. Neumann $et~al.$~\cite{neumann2005color} introduced global parameter control with histogram optimization to improve the accuracy of color transfer. Ma $et~al.$~\cite{ma2009color} build a nonlinear color map with neighborhood analysis for color transfer. Pouli $et~al.$~\cite{pouli2011progressive} used high dynamic range control in color transfer to generate more natural result. Rabin $et~al.$~\cite{rabin2014adaptive} utilized optimal transfer to map global color distributions between images. Su $et~al.$~\cite{su2014corruptive} proposed a self-learning filtering scheme to extract semantic-based correspondences in color transfer. Han $et~al.$~\cite{han2016cartoon} took full use of image decomposition to improve the performance of color transfer. Kaur $et~al.$~\cite{kaur2020color} proposed an efficient gradient channel prior to guide color correction for dehazing. Yan $et~al.$~\cite{yan2021qhsl} designed a color model that is used to define a representation of two-dimensional QHSL images for efficient color transformation. In summary, the global color transfer methods can be implemented with more simple strategy. It doesn't require complex semantic analysis that increases the practicality. However, the performance of the methods depends on the similarity of global color distributions between images. In addition, the methods cannot generate semantic-based color transfer results according to different objects in image.

The correspondence-based color transfer methods implement color mapping based on palette or semantic correspondence. Tai $et~al.$~\cite{tai2005local} proposed a Gaussian mixture model (GMM)-based image segmentation to generate more accurate color transfer. Abadpour $et~al.$~\cite{abadpour2007efficient} used principal component analysis to cluster colors for color mapping. Murray $et~al.$~\cite{murray2012toward} proposed a semantic-aware color transfer method to improve the correspondence of color distributions. Chang $et~al.$~\cite{chang2015palette} proposed a palette-based recoloring method to preserve inherent color characteristics. Iwasa $et~al.$~\cite{iwasa2018color} proposed a method to extract palette colors and edit them in image. Liu $et~al.$~\cite{liu2021palette} utilized intrinsic decomposition to optimize illumination for palette-based image recoloring. HaCohen $et~al.$~\cite{hacohen2011non} proposed a patchmatch-based method to connect non-rigid areas in images to establish color mapping. Arbelot $et~al.$~\cite{arbelot2017local} presented a heuristic texture descriptor to detect correspondence for color transfer. Yamamotot $et~al.$~\cite{yamamoto2007colour} detected SIFT-based correspondence between images to guide color transfer. Tehrani $et~al.$~\cite{tehrani2010iterative} improved the framework for multi-view images. Park $et~al.$~\cite{park2016efficient} used low-rank matrix factorization to provide global parameter control based on SIFT-based correspondence. It achieves better illumination in color transfer result. Such methods detect dense or sparse correspondence to implement more accurate color transfer. However, the robustness of the correspondence detection is affected by various influencing factors, including exposure, shadow, different poses, blend texture, etc.

The deep feature-based color transfer methods represent the latest trend in the field. Benefited from the robust semantic analysis and feature learning on large dataset, more accurate color transfer result can be generated based on the trained deep neural network. He $et~al.$~\cite{he2018deep} proposed a CNN-based color transfer framework with multi-source combination function. It achieves better performance for semantic analysis. He improved the framework with the extend semantic detection~\cite{he2019progressive} in lower feature layers. Zhang $et~al.$~\cite{zhang2019deep} utilized the similar structure to implement color transfer between videos. The inter-frame similarity is well studied to improve the semantic consistency. Lee $et~al.$~\cite{lee2020deep} combined CNN-based feature analysis and histogram optimization to avoid discontinuities in color distribution. Huang $et~al.$~\cite{huang2020learning} proposed a generative adversarial network to implement High-Dynamic-Range image color transfer. Jiang $et~al.$~\cite{jiang2021cotr} proposed the lighting robust framework to detect correspondences based on transformer. Guo $et~al.$~\cite{guo2020zero} designed a tone curve estimation network to adjust color distribution for color correction. Liu $et~al.$~\cite{liu2021retinex} presented a low-light enhancement framework by retinex-based neural architecture searching model to improve the quality of color distribution. Ding $et~al.$~\cite{ding2022deep} proposed a style transfer method that utilizes wavelet transformation for producing content-preserving stylized results. Shao $et~al.$~\cite{shao2023hairstyle} presented a hairstyle color transfer method based on StyleGAN2. At the same time, color distributions are transferred with better semantic correspondence. The mentioned methods take full advantages of feature analysis and large dataset-based color distribution fitting by deep learning structure to generate high quality color scheme. However, the methods still can't effectively transfer colors between images without semantic correspondence.

We propose a new palette-based color transfer framework that establishes color mapping according to palette-based clustering and image segmentation. Our palette generation is more accurate and robust to images with scattered color distribution. It doesn't require user-specified cluster centers. For color mapping, our framework doesn't require the strict semantic correspondence between source image and reference one. Benefited from a deep learning-based image segmentation, the mapping is implemented for foreground and background parts independently. It achieves semantically irrelevant color transfer strategy while keeping the naturalness of the transfer result. In following parts, we introduce the details of our framework.

\section{Palette-based Clustering}

For palette-based color transfer or recoloring works, the most important issue is how to generate a palette to represent color distribution. One classical solution is to cluster colors into different categories with a user-specified cluster center number~\cite{tai2005local}\cite{chang2015palette}. However, different center numbers can lead to different palettes for the same image, resulting in unpredictable effects for color mapping. For example, if the user selects a smaller center number for an image with a complex color distribution, a large number of unrelated colors may be grouped into one category, leading to incorrect color transfer results. An ideal palette generation scheme should consider various color distributions for different images and generate adaptive palettes. Similar colors should be classified into the same item of the palette while maintaining separation for different colors. Based on these requirements, we propose the palette-based clustering method.

\begin{figure}
  \centering
  \includegraphics[width=\linewidth]{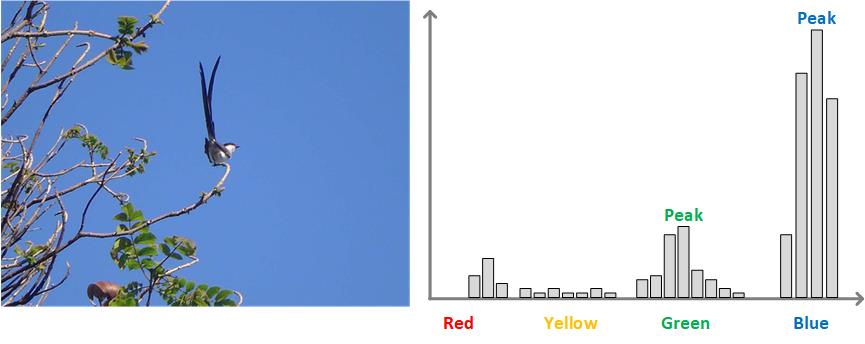}
  \caption{An instance of peak values in histogram.}
  \label{f1_1}
\end{figure}

The clustering method establishes a palette for the image based on color histogram analysis in the Lab color space. The color histogram reflects aggregate information for different colors. For instance, as shown in Figure~\ref{f1_1}, a landscape image with the sky part contains a peak in the blue range of the histogram. The clustering method searches for such peak values to generate a palette while aggregating related adjacent colors. The searching process is implemented in the Lab color space because the color representations in the RGB space are coupled across three channels. {Additionally, the influence of various lighting conditions cannot be conveniently removed.} In the Lab color space, the color representations in different channels are decoupled, making it easier to search for accurate peak values and eliminate lighting effects. {Based on these principles, we provide implementation details of our palette-based clustering method, which includes three core parts: \textbf{histogram construction, peak searching,} and \textbf{peak merging}.} 

For \textbf{histogram construction}, we divide each channel $(l_i,a_i,b_i)$ of Lab into specified $z$ bins and accumulate pixels into the related bin $b_i$ ($z$ = 100 by default). Then, a continuous color distribution is converted to a discrete form for different channels. This process is shown as
\begin{equation}
\label{e1}
I_H = \{b_0,...,b_z\},b_i=(l_i,a_i,b_i),i\in[0,z],
\end{equation}
where $I_H$ represents the histogram of input image $I$ in Lab color space. For \textbf{peak searching}, we compare each bin $b_i$ one by one and select the peak that has maximum value in a small range $N\{b_i\}$ into the candidate peak set $\{b_p\}$. The formulation can be represented as:
\begin{equation}
\label{e2}
\begin{array}{c}
\{b_p\} = \max\{{b_i}|N({b_i}), b_i>b_{min}\},\\
N\{b_i\} = \{b_{i-r},...,b_{i+r}\}
\end{array}
\end{equation}
{where $r$ represents the searching radius, which defines the scale of local range $N\{b_i\}$.} In practice, we set $r$ to 3 as a default value. It should be noticed that the peak searching is implemented in local range. Random color distributions may produce some fake peak values without statistical significance. Such peak values affect the accurate of palette generation. To reduce the influence, we add a threshold-based control ($b_i>b_{min}, b_{min}=30$ by default) to avoid adding some peak with few related pixels. Based on the mentioned implementation details, we obtain the candidate peak set $\{b_p\}$.

\begin{figure}
  \centering
  \includegraphics[width=\linewidth]{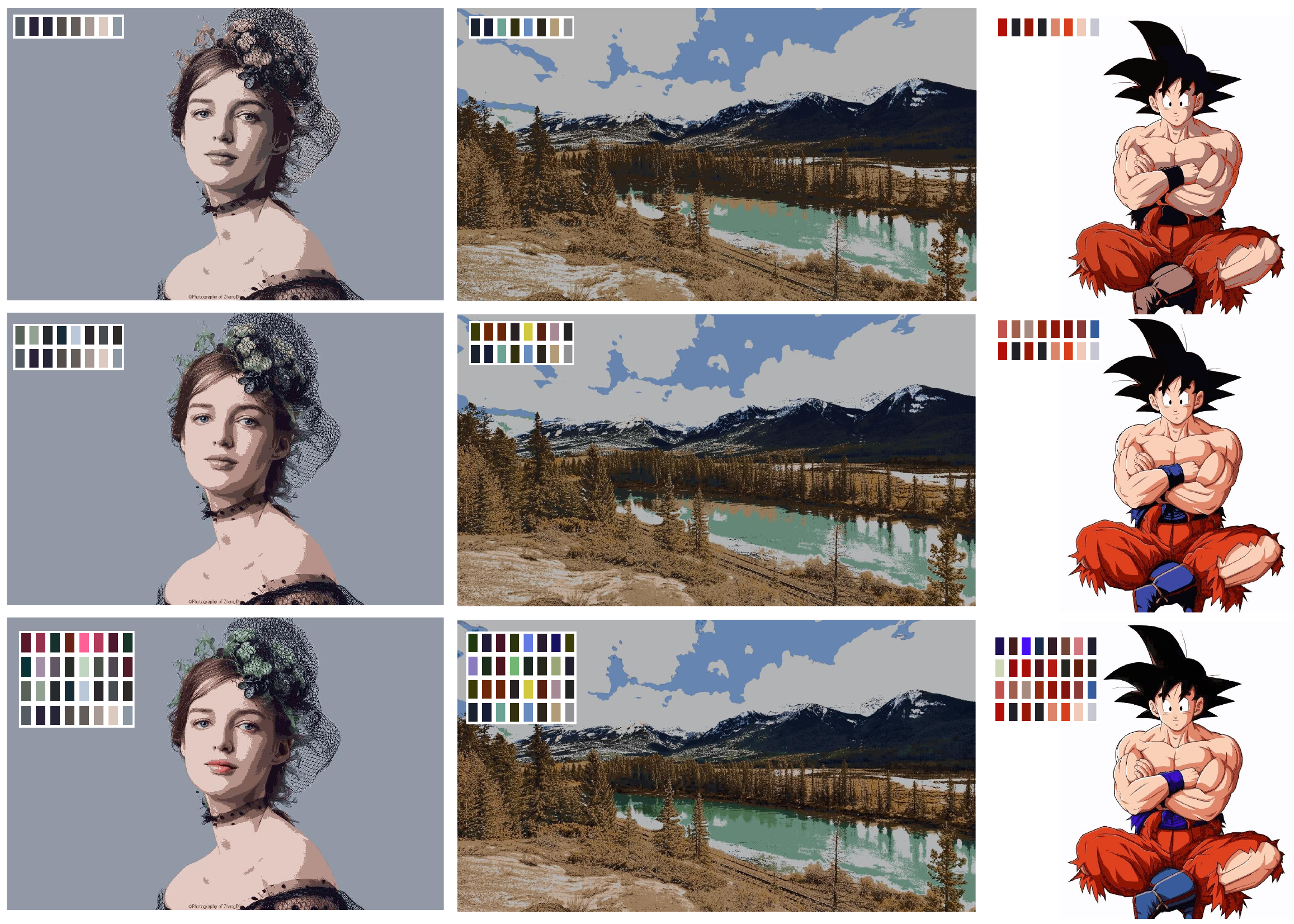}
  \caption{Some instances of palette-based clustering with different upper bound numbers (8,16,32). All pixels are revalued by their related palette values.}
  \label{f2}
\end{figure}

Theoretically, we can compute the palette from the ${b_p}$ directly. {It is similar to the center-based clustering strategy that clusters different colors into related palette items according to specified centers.} In our framework, the centers are replaced by the peaks. {However, we have observed that the scale of the peak set ${b_p}$ exceeds the acceptable range in most cases, resulting in a generated palette that is too large, thus reducing the aggregation property for colors.} To address this issue, we introduce a peak merging step to reduce the scale of ${b_p}$. We collect the achieved ${b_p}$ from the three channels of the Lab color space. By combining the three channels, we generate one set of peak colors and construct a kd-tree from the set. Then, we input the Lab value of each pixel one by one into the kd-tree and count the number related to each peak. An upper bound number $t$ for the peak set can be specified. The final peak set ${b_{fp}}$ is computed by
\begin{equation}
\label{e3}
\{{b_{fp}}\}  = \{ {b_i}|K(b_i) > K({b_t}),{b_i} \in \{ {b_p}\},\}
\end{equation}
where $K$ represents the pixel number accumulation for related bin $b_i$ by kd-tree searching, $b_t$ is the bin that the order of $K({b_t})$ is same to $t$ in $K\{b_p\}$. Once the final peak set $\{b_{fp}\}$ is achieved, all pixels can be classified based on the set. The classification is shown as 
\begin{equation}
\label{e4}
{p_i} = \arg\min d(L({p_i}),b_i),b_i \in \{ {b_{fp}}\}
\end{equation}
{where $p_i$ represents a pixel of the input image $I$, $L(p_i)$ represents the Lab values of $p_i$, and $d$ is the distance between $L(p_i)$ and different peak values in ${b_{fp}}$.} The classification result of $p_i$ is the index of $b_i$ which has smaller value of $d$ in Lab color space. Based on the equation, the pixels are classified into different categories. The palette can be generated based on the average values in related categories. The selection of upper bound number $t$ controls the scale of final peak set that decides the length of palette. In Figure~\ref{f2}, we show some instances of palette-based clustering with different values of $t$. To balance the aggregation property and scale control of palette, we set $t$ to 32 that is based on the experience. It guarantees that most of the color information can be preserved into the palette.

\section{Color Mapping Strategy}

{Based on the palette-based clustering, the color transfer task can be conveniently  transformed to find color mapping $f$, $f:I_s\{b_{fp}\}\to I_r\{b_{fp}\}$.} $I_s$ represents the source image and $I_r$ represents the reference one. In traditional works of palette-based recoloring, the mapping $f$ is specified by user. In our framework, we aim to implement $f$ as an automatic processing. To achieve the goal, we propose three constraints to build the mapping $f$, including \textbf{split correspondence, chromatic aberration control,} and \textbf{color consistency keeping}. 

The \textbf{split correspondence} means that the mapping is implemented for the foreground and background parts independently. For most images, the internal content can be divided into two parts: foreground and background. {The foreground typically represents semantic objects such as people, animals, and buildings, while the background serves as a context or backdrop to highlight the foreground objects.} It is natural to implement color mapping for the foreground and background separately. With advancements in deep learning frameworks, the separation between foreground and background parts can be accurately achieved. In our framework, we utilize DeepLabV3+~\cite{chen2018encoder} to achieve this separation. Based on this separation, the color mapping function $f$ is divided into two parts:
\begin{equation}
\label{e5}
f = \left\{\begin{array}{*{20}{c}}
{{f_{fore}}:{I_s}{{\{ {b_{fp}}\} }_{fore}} \to {I_r}{{\{ {b_{fp}}\} }_{fore}}}\\
{{f_{back}}:{I_s}{{\{ {b_{fp}}\} }_{back}} \to {I_r}{{\{ {b_{fp}}\} }_{back}}}
\end{array},\right.
\end{equation}
where $f_{fore}$ and $f_{back}$ represent the color mapping for the foreground and background parts, $\{b_{fp}\}_{fore}$ and $\{b_{fp}\}_{back}$ represent palette-based categories belong to the related parts. The split correspondence keeps the basic correspondence for foreground and background color distributions. The split correspondence is independent to the specific semantic information between images. It doesn't require the strict semantic correspondence to implement color mapping. Once the images cannot be divided into clear foreground and background, all pixels in images are considered as the foreground, which keeps the method has adequate generality.

\begin{figure}
  \centering
  \includegraphics[width=\linewidth]{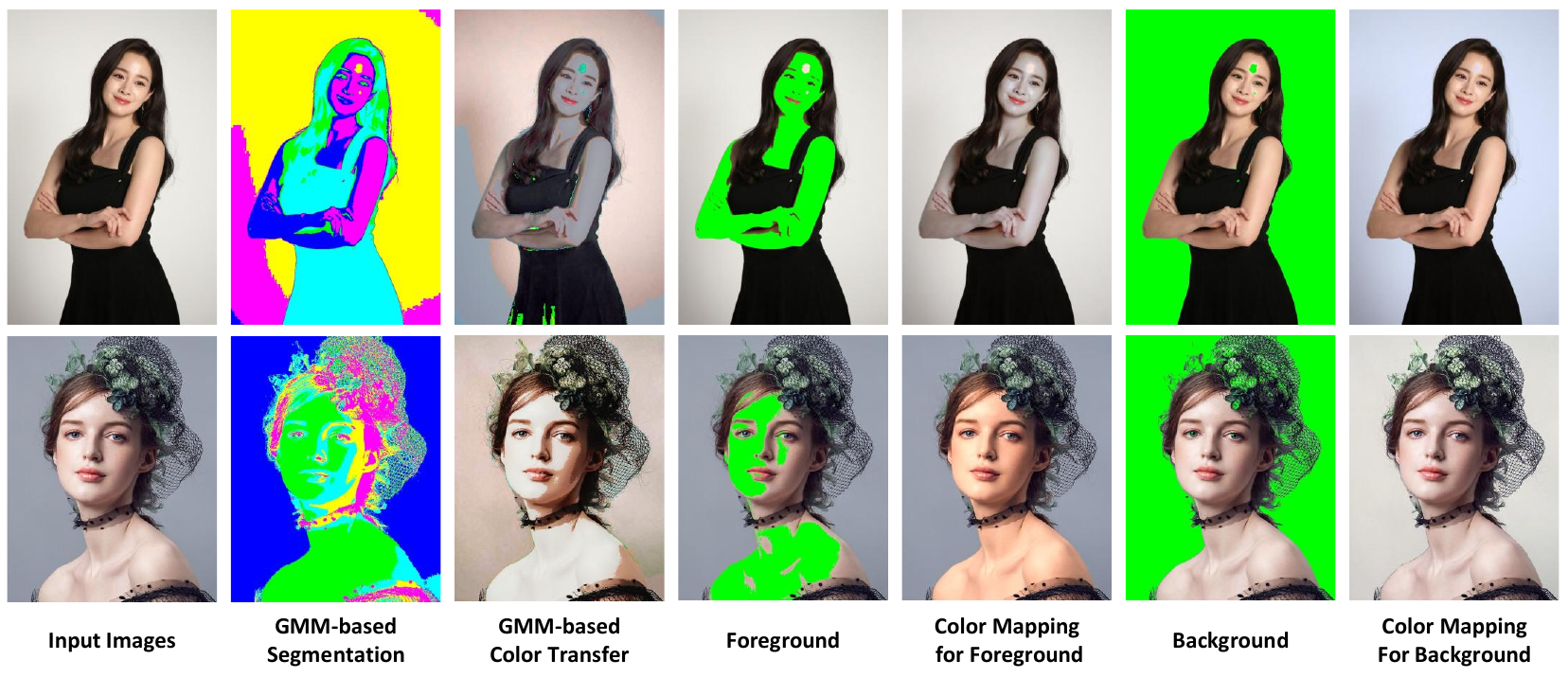}
  \caption{Comparisons between GMM-based color transfer strategy and palette-based one. First column: input images for color transfer to each other; second column: GMM-based segmentation results(the colors have no correspondence between images); third column: GMM-based color transfer results\cite{tai2005local}; other columns: split correspondence for palette-based color mapping.}
  \label{f4}
\end{figure}

The \textbf{chromatic aberration control} means that the color mapping between different peak values of related palettes should be controlled in a reasonable range. For instance, a random mapping may change colors of regions with high light representation to low one. Such mapping breaks light intensity-based semantic representation. \textbf{red}{In another condition, the palette-based color mapping may not change the color distribution of the source image based on ambiguous correspondence.} To control the mapping, we use the nearest neighbor searching to find the $b_j$ from ${I_r}{\{{b_{fp}}\}}$ for $b_i$, represented as
\begin{equation}
\label{e6}
{f_i}:{b_i} \to {b_j},{b_j} = \arg \min d({b_i},{b_j}),
\end{equation}
where $d$ is the distance in Lab color space which has been used in Equation~\ref{e4}. Based on the nearest neighbor searching, the basic information of chromatic aberration can be inherited from source image to transfer result while considering the color scheme of reference one. Although the nearest neighbor searching strategy for chromatic aberration control reduces the flexibility of color mapping, it avoids disorder color transfer. The property is important for the generality in practice. In Figure~\ref{f4}, two instances of color mapping with split correspondence and chromatic aberration control are shown. It is clear that the quality of mapping is improved based on the two constraints.

{The \textbf{color consistency keeping} aims to maintain the basic structure of color distribution during the color mapping process. A negative example is discussed in Figure \ref{f4}.} The GMM-based color classification fails to accurately capture palette information with a specified center number. The resulting inaccurate classification (shown in the second column of Figure \ref{f4}) disrupts the basic structure of color distribution, leading to unsatisfactory transfer results (shown in the third column of Figure \ref{f4}). In our framework, we introduce a color consistency keeping method to preserve the structure of color distribution. This method consists of internal consistency keeping and external continuity keeping.

{In the part of chromatic aberration control, we discussed the use of nearest neighbor searching to map peak values from the reference image to the source image. However, this searching strategy does not guarantee a one-to-one mapping, meaning that some peak values of the source image may share the same peak value from the reference image. When a many-to-one mapping is established, the structure of color distribution is inevitably disrupted.} It is represented as
\begin{equation}
\label{e7}
{f_I}:\{ {b_I}\}  \to {b_j},{b_i} \in \{ {b_I}\},
\end{equation}
where $\{b_I\}$ is a subset from source image and it shares a same peak value $b_j$ of reference image. The structure represented by $\{b_I\}$ in source image is compressed after mapping. To solve the problem, the internal consistency keeping is proposed. The related formulation is represented as
\begin{equation}
\label{e7}
{f_k}:b_k \to b_j,b_k=max\{b_i|b_i\in\{b_I\}\},
\end{equation}
where $f_k$ is used to instead $f_I$ and $b_k$ is selected from $\{b_I\}$ with the maximum number of pixels. Other peak values in $\{b_I\}$ are added into the pending set $\{b_Q\}$. Then, the many-to-one mapping is eliminated.

The external continuity keeping is to finish the color mapping for the set $\{b_Q\}$. The mapping values of $\{b_Q\}$ are computed based on existing mapping results with corresponding weights. The mapping can be formulated as
\begin{equation}
\label{e8}
{f_q}:{b_q} \to \sum\limits_{{b_k} \in N({b_q})} {{w_k}f({b_k})},{b_q} \in \{{b_Q}\},
\end{equation}
\begin{equation}
\label{e9}
{w_k} = \frac{{d{{({b_k},{b_q})}^{ - 1}}}}{{\sum\limits_{{b_k} \in N({b_q})} {d{{({b_k},{b_q})}^{ - 1}}} }},
\end{equation}
where $b_p$ is a peak value of $\{b_P\}$, $b_k$ represents the peak value which has obtained mapping value, $N({b_p})$ is a neighbor set of $b_P$. The mapping value of $b_p$ is computed by its neighbor’s mapping. The related weight is computed by the reciprocal of distance $d$. Combining the implementations of split correspondence, chromatic aberration control, and color consistency keeping, our color mapping strategy is established. For each pixel in source image, the related color transfer result can be computed by
\begin{equation}
\label{e10}
f(p_i) = f(b_i) + L(p_i) - b_i,
\end{equation}
where $p_i$ is a pixel in source image $I_s$, $b_p$ is the peak value related to the $p$ in palette, $L(p)$ is the Lab values of $p$, $p_i'$ is the related transfer result, $p_i' = f(p_i)$. Based on the color mapping result $f(p_i)$, the new Lab-based values $L(p_i')_a$ and $L(p_i')_b$ in a and b channels of $p_i'$ can be achieved by $L(p_i')_a = f(p_i)_a,L(p_i')_b = f(p_i)_b$. It should be noticed that the value $L(p')_l$ of $L$ channel of $p$ is not changed in practice. We update $L(p')_l$ in lighting optimization.

\begin{figure}
  \centering
  \includegraphics[width=0.8\linewidth]{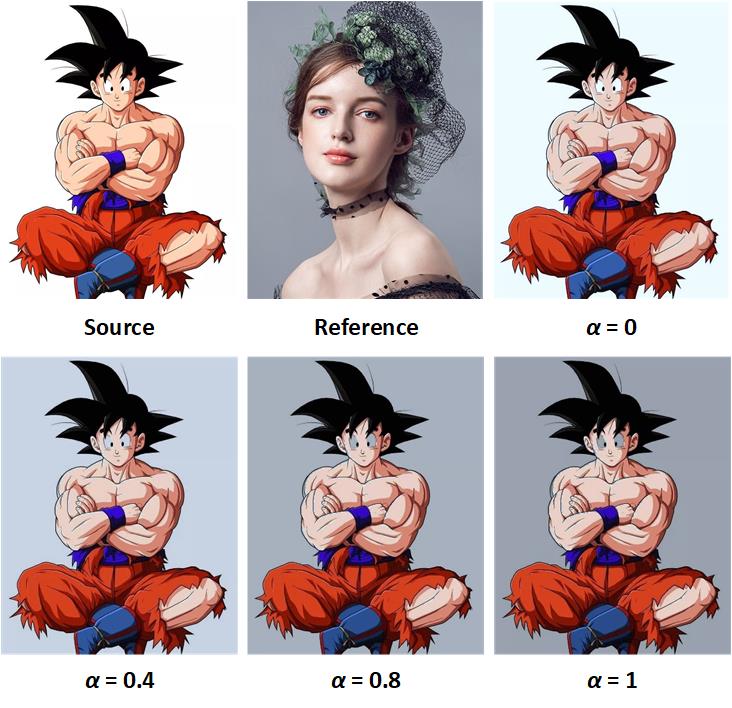}
  \caption{An instance of color mapping-based weighted update for $L$ value with different weights.}
  \label{f5}
\end{figure}

\section{Lighting Optimization}

After color mapping, the values of the $a$ and $b$ channels for the source image have been transferred according to those of the reference image. {However, the values in the $L$ channel are processed individually from the mapping. The reason is that such values reflect the lighting intensity of the pixels. Despite the introduction of various color control constraints in the color mapping strategy, it still involves discrete computation, which carries the potential risk of producing discontinuous color distribution.} Furthermore, the exposure quality of the reference image directly affects the quality of color transfer. If the values in the $L$ channel are updated using Equation~\ref{e10}, any abnormal exposure in the reference image will be transferred into the mapping result simultaneously. Therefore, the values in the $L$ channel should be optimized separately in an independent step.

We design a lighting optimization to be the step. It is constructed by two part: color mapping-based weighted update and global lighting enhancement. The color mapping-based weighted update is to add mapping result from $f(p_i)$ to the original $L$ channel with a specified weight, represented as
\begin{equation}
\label{e12}
L{(p_i')_l} = (1-\alpha) L{(p_i)_l} + \alpha f{(p_i)_l},
\end{equation}
where $\alpha$ is used to control the weight of $f(p_i)_l$ that is the color mapping result in $L$ channel. In Figure \ref{f5}, we show the influence of different weights in color transfer. In practice, we set $\alpha$ to 0.3 by default. 

\begin{figure}
  \centering
  \includegraphics[width=0.8\linewidth]{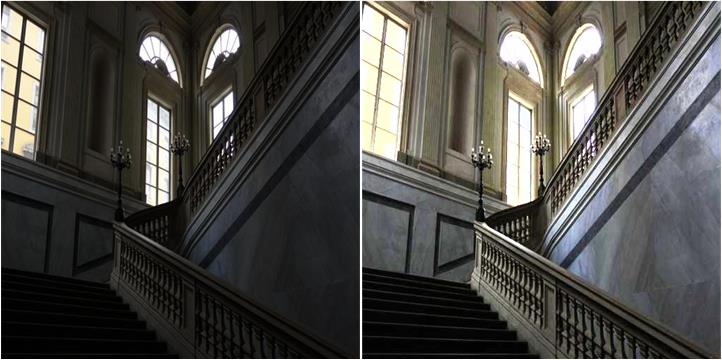}
  \caption{An instance of global lighting enhancement.}
  \label{f6}
\end{figure}

{For some reference images with abnormal exposure regions, we optionally employ the global lighting enhancement to optimize the values in the $L$ channel.} The basic framework of this enhancement is implemented according to~\cite{ma2022toward}. We extract the values of the $L$ channel processed by the enhancement and reassign them as $L{(p')_l}$. This process helps improve the quality of transfer results with abnormal exposure regions produced by the reference image. An example of global lighting enhancement is depicted in Figure \ref{f6}. In experiments, we demonstrate the performance of the proposed color transfer framework."

\begin{figure*}[ht]
  \centering
  \includegraphics[width=0.85\linewidth]{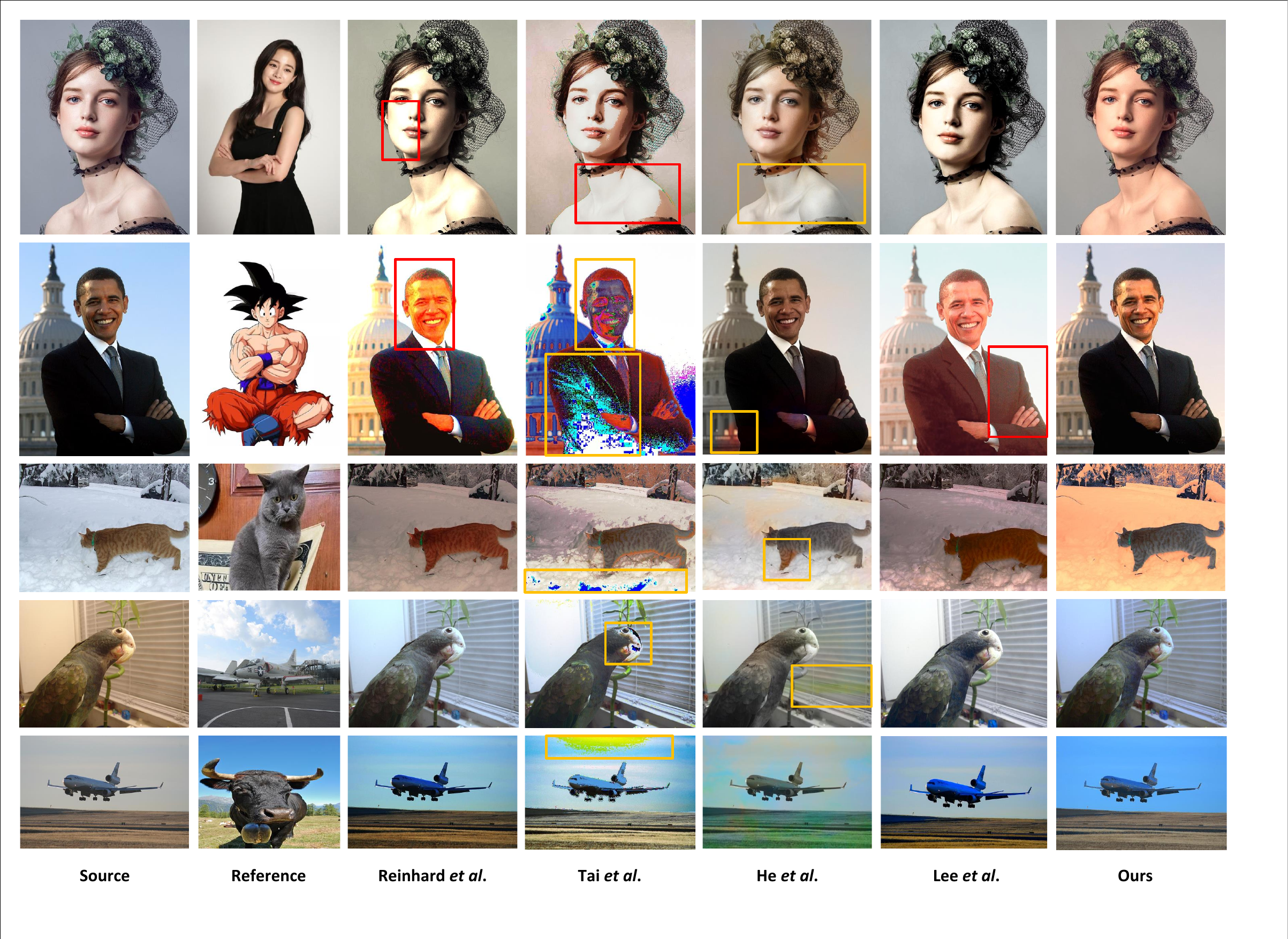}
  \caption{Comparisons between different color transfer methods. Methods: Reinhard $et~al.$~\cite{reinhard2001color}, Tai $et~al.$~\cite{tai2005local}, He $et~al.$~\cite{he2018deep}, Lee $et~al.$~\cite{lee2020deep}. It is clear that our method achieves more natural transfer results for different images. Red boxes: abnormal exposure regions; yellow boxes: color misalignment or degradation regions.}
  \label{fe1}
\end{figure*}

\section{Experiments}

In this section, we report the performance of our color transfer framework for various pictures. The test dataset is collected from three sources: VOC2012~\cite{voc2012}, Zero-DCE dataset~\cite{guo2020zero}, and random selections from the internet. The images in the dataset are classified into two groups: separable pictures and inseparable ones. The separable picture means that the foreground and background parts can be detected independently. The inseparable picture is just the opposite. The hardware platform is constructed with Intel Xeon W 2133 3.6G Hz(CPU), 32 GB RAM, and Quadro P620(GPU). The development platform is built in Visual Studio 2019 (64 bit) on Windows 11. Except for some visual comparison results, we design two quantitative analysis tools, color consistency analysis and fading rate computation, to measure the performance of color transfer methods. We also introduce some image quality assessments to evaluate the quality of the methods. The assessments includes Brisque~\cite{mittal2012no}, ClipIQA+~\cite{wang2022exploring}, CNNIQA~\cite{kang2014convolutional}, DBCNN~\cite{zhang2018blind}, MUSIQ~\cite{ke2021musiq}, and NIMA~\cite{talebi2018nima}. Finally, we discuss some limitations of our framework.

\begin{figure}[h!t]
  \centering
  \includegraphics[width=\linewidth]{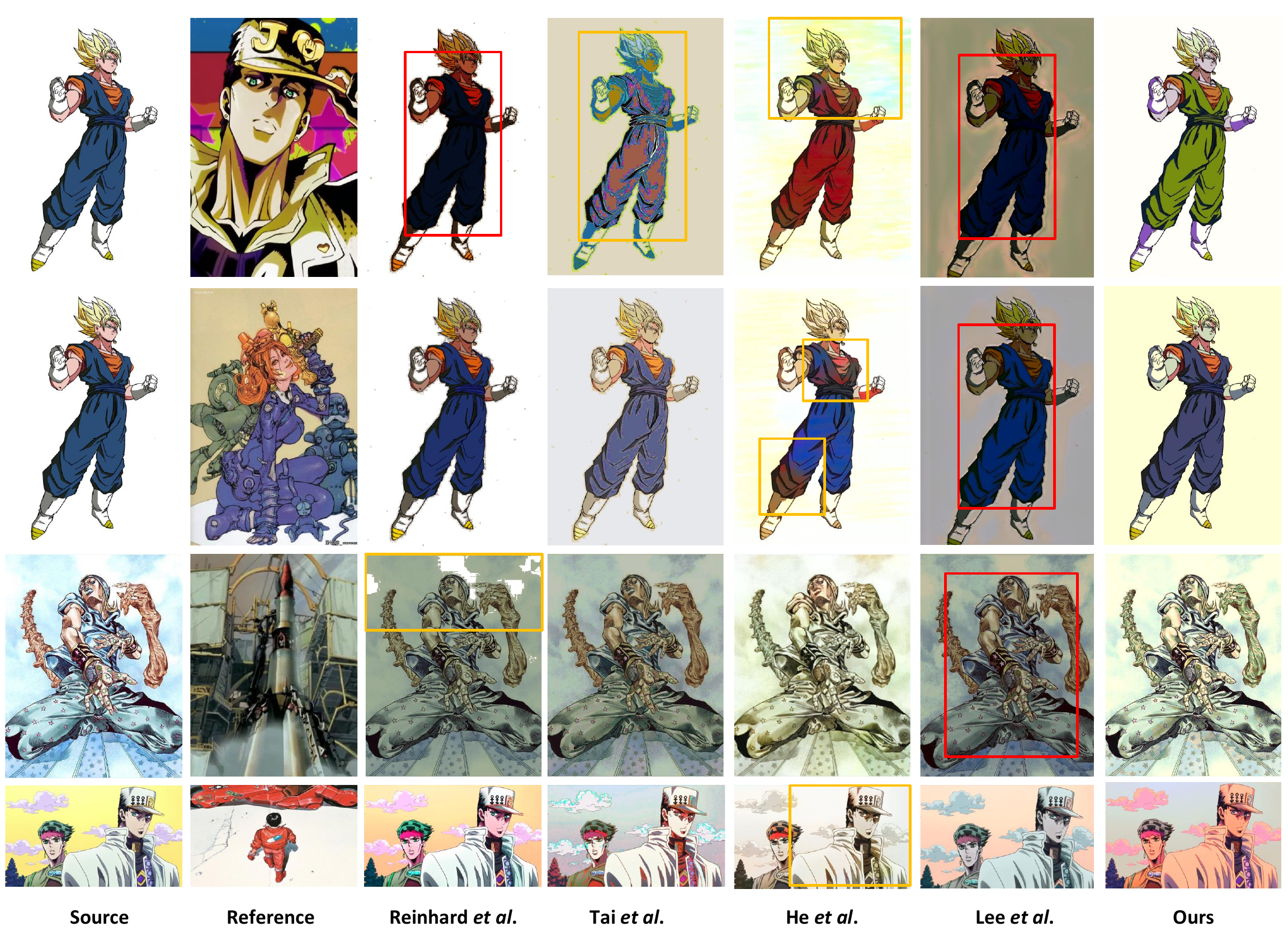}
  \caption{Comparisons between different color transfer methods for comically pictures. Our method obtains better exposure results while fitting the color distributions according to foreground and background of reference. Red boxes: abnormal exposure regions; yellow boxes: color misalignment or degradation regions.}
  \label{fe2}
\end{figure}

\subsection{Color Transfer on Separable Pictures}

As mentioned before, the separable pictures have detachable foreground and background parts. The human visual perception is more sensitive to the foreground parts that are often regarded as the saliency region~\cite{yang2019sgdnet}. Naturally, the color transfer should consider such separability to achieve more subjective results. We compare different color transfer methods for performance demonstration on separable pictures. The methods include Lab color transfer~\cite{reinhard2001color}, GMM-based color transfer~\cite{tai2005local}, deep exemplar-based color transfer~\cite{he2018deep}, and deep histogram analysis-based color transfer~\cite{lee2020deep}. Some semantic correspondence methods are excluded from the comparison. The reason is that the generality of such methods is poor for images without close semantic correspondence. In Figure~\ref{fe1}, some comparison results are shown. The Lab color transfer constructs global color mapping between images without independent processing for foreground and background parts. It cannot achieve accurate transfer result for foreground objects (cat picture in the third row). The performance of GMM-based color transfer is limited by the accuracy of segmentation, leading to color overflow produced by incorrect color segmentation (as observed in the Obama portrait photo in the second row). Some color fading and abnormal exposure regions are produced by deep learning methods. In contrast, our method achieves more stable and accurate color transfer results. We will report more details on larger test set in quantitative analysis.

\begin{table}[t]
    \caption{
        Color consistency and degradation estimation on images shown in Figures \ref{fe1} and \ref{fe2}.
    }
    \label{T1_internet}
    \centering
    \setlength{\tabcolsep}{3pt}
    \resizebox{\linewidth}{!}{
    \begin{tabular}{lcccccc}
        \toprule
        \multicolumn{1}{l}{} && \multicolumn{2}{c}{\textbf{Consistency}}  && \multicolumn{2}{c}{\textbf{Fading Rate}}\\
        \cmidrule(lr){3-4} \cmidrule(lr){6-7}
        \textbf{Method} && \textbf{L} & \textbf{RGB} && \textbf{F-a} & \textbf{F-b} \\
        \midrule
        \textbf{Reinhard $et~al.$~\cite{reinhard2001color}} && 9.05E-03  & 7.73E+02  && 2.50E-02 &  \textbf{4.76E-03}  \\
        \textbf{Tai $et~al.$~\cite{tai2005local}} && 3.18E-02  & 1.41E+03  && 5.65E-02 &  7.11E-03\\
        \textbf{He $et~al.$~\cite{he2018deep}}    && 2.19E-01  & 3.45E+03  && 2.74E-02 &  6.17E-03 \\
        \textbf{Lee $et~al.$~\cite{lee2020deep}} && 3.71E-01  & 2.28E+03  && 4.50E-02 &  7.71E-03\\
        \midrule
        \midrule
        \textbf{Ours}   && \textbf{5.48E-03}  & \textbf{2.52E+02}  && \textbf{2.34E-02} &  5.75E-03 \\        
        \bottomrule
    \end{tabular}
    }
\end{table} 

\begin{table}[]
\centering
\caption{Image quality assessment with different metrics on images shown in Figures \ref{fe1} and \ref{fe2}.}
\setlength{\tabcolsep}{3pt}    
\resizebox{\columnwidth}{!}{
\begin{tabular}{lcccccc}
\toprule 
Methods        & \textbf{Brisque} & \textbf{Clipiqa+} & \textbf{CNNIQA} & \textbf{DBCNN}   & \textbf{MUSIQ}   & \textbf{NIMA}   \\ \midrule
\textbf{Reinhard $et~al.$~\cite{reinhard2001color}}            & 24.8992          & 0.5631            & 0.7003          & 57.7676          & 65.3644          & 4.8663          \\
\textbf{Tai $et~al.$~\cite{tai2005local}}           & 25.3682          & 0.4694            & 0.6086          & 55.8637          & 61.2941          & 4.4928          \\
\textbf{He $et~al.$~\cite{he2018deep}} & 23.0492          & 0.5781            & 0.5131          & 59.8861          & 63.7291          & 4.9301          \\
\textbf{Lee $et~al.$~\cite{lee2020deep}}      & 20.1237          & 0.5519            & 0.6426          & 60.2026          & 64.3713          & 4.9043   \\ 
\midrule
\midrule
\textbf{Ours}           & \textbf{26.4261} & \textbf{0.5937}   & \textbf{0.7124} & \textbf{62.6177} & \textbf{66.4929} & \textbf{4.9361} \\ \bottomrule
\end{tabular}}
\label{T2_internet}
\end{table}

\begin{figure*}[ht]
  \centering
  \includegraphics[width=\linewidth]{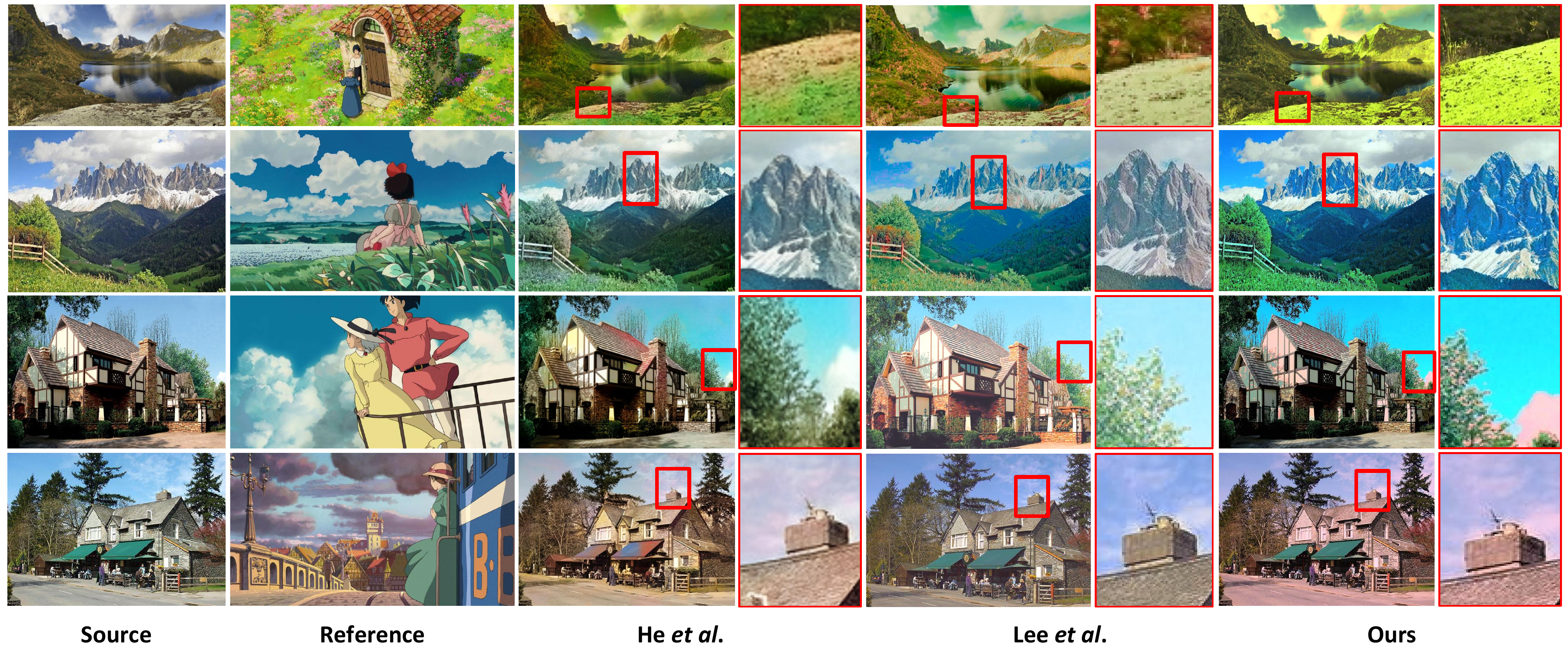}
  \caption{Comparisons between deep-learning frameworks and our method for landscape pictures. Our method achieves similar quality transfer results and enhances some color distributions according to the references (first and second rows, the green and blue signals are enhanced). }
  \label{fe3}
\end{figure*}

\subsection{Color Transfer on Inseparable Pictures}

It should be noticed that some inseparable pictures cannot be divided into distinct foreground and background parts. The color transfer method should be able to implement color mapping for the pictures. We select some cartoon pictures with uncertain separability to show the color transfer results. In Figure~\ref{fe2}, the results by different methods are shown. For separable cartoon pictures, our method can achieve accurate results just like in Figure~\ref{fe1}. With some inseparable reference pictures (first and third rows), our method can also achieve the color transfer result that keeps the consistency of global color distribution. For some inseparable pictures with more complex color distributions, the advantage of split correspondence in our color mapping strategy is disappear. In 
such cases, our method can still obtain acceptable color transfer results. Benefited from the chromatic aberration control, color consistency keeping, and lighting optimization, the performance of our method can be kept. In Figure~\ref{fe3}, we compare the SOTA methods with our frameworks on landscape images with the scattered color distributions. It is clear that our method avoids color degradation while keeping the global color distributions from reference images to the transfer results. More details are reported in following part.

\begin{table}[t]
    \caption{
        Color consistency and degradation estimation in VOC2012 test set. Ours(R) means that the lighting optimization is removed from our framework.
    }
    \label{T3_VOC}
    \centering
    \setlength{\tabcolsep}{3pt}
    \resizebox{\linewidth}{!} {
    \begin{tabular}{lcccccc}
        \toprule
        \multicolumn{1}{l}{} && \multicolumn{2}{c}{\textbf{Consistency}}  && \multicolumn{2}{c}{\textbf{Fading Rate}}\\
        \cmidrule(lr){3-4} \cmidrule(lr){6-7}
        \textbf{Method} && \textbf{L} & \textbf{RGB} && \textbf{F-a} & \textbf{F-b} \\
        \midrule
        \textbf{Reinhard $et~al.$~\cite{reinhard2001color}} && \textbf{5.36E-03}  & 8.18E+02  && 2.39E-02 &  \textbf{5.76E-03}  \\
        \textbf{Tai $et~al.$~\cite{tai2005local}} && 1.79E-02  & 1.26E+03  && 4.98E-02 &  1.61E-02\\
        \textbf{He $et~al.$~\cite{he2018deep}}    && 1.66E-01  & 4.86E+03  && 2.71E-02 &  6.18E-03 \\
        \textbf{Lee $et~al.$~\cite{lee2020deep}} && 3.43E-01  & 7.23E+02  && 2.84E-02 &  5.41E-03\\
        \midrule
        \midrule
        \textbf{Ours(R)}   && 2.42E-02  & 4.54E+02  && 3.42E-02 &  8.77E-03 \\ 
        \textbf{Ours}   && 1.47E-02  & \textbf{4.41E+02} && \textbf{2.18E-02} &  7.05E-03 \\ 
        \bottomrule
    \end{tabular}
    }
\end{table} 

\begin{table}[t]
\centering
\caption{Image quality assessment with different metrics on images of VOC2012.}
\setlength{\tabcolsep}{3pt}    
\resizebox{\columnwidth}{!}{
\begin{tabular}{lcccccc}
\toprule 
Methods         & \textbf{Brisque} & \textbf{Clipiqa+} & \textbf{CNNIQA} & \textbf{DBCNN} & \textbf{MUSIQ} & \textbf{NIMA} \\ \midrule
\textbf{Reinhard $et~al.$~\cite{reinhard2001color}} & 23.32            & 0.5953            & 0.8212          & 61.27          & 67.02          & 4.371         \\
\textbf{Tai $et~al.$~\cite{tai2005local}}      & 27.68           & 0.4881            & \textbf{0.9092}          & 63.16          & 65.93          & 4.402         \\
\textbf{He $et~al.$~\cite{he2018deep}}       & 15.09            & \textbf{0.6196}            & 0.0267          & 61.14          & 64.76          & \textbf{4.863}         \\
\textbf{Lee $et~al.$~\cite{lee2020deep}}      & 20.32            & 0.6133            & 0.8194          & 62.81          & 67.84         & 4.463         \\
\midrule
\midrule
\textbf{Ours(R)}     & 24.68            & 0.5716            & 0.7713          & \textbf{63.38}          & 67.26          & 4.402  \\
\textbf{Ours}     & \textbf{28.93}           & 0.5741            & 0.7651          & 63.06        & \textbf{68.11}          & 4.412 \\

\bottomrule      
\end{tabular}}
\label{T4_VOC}
\end{table}

\begin{table}[t]
    \caption{
        Color consistency and degradation estimation in VOC2012 test set reference by Zero-DCE images.
    }
    \label{T5_DCE}
    \centering
    \setlength{\tabcolsep}{3pt}
    \resizebox{\linewidth}{!}{
    \begin{tabular}{lcccccc}
        \toprule
        \multicolumn{1}{l}{} && \multicolumn{2}{c}{\textbf{Consistency}}  && \multicolumn{2}{c}{\textbf{Fading Rate}}\\
        \cmidrule(lr){3-4} \cmidrule(lr){6-7}
        \textbf{Method} && \textbf{L} & \textbf{RGB} && \textbf{F-a} & \textbf{F-b} \\
        \midrule
        \textbf{Reinhard $et~al.$~\cite{reinhard2001color}} && 5.24E-03  & 1.19E+03  && \textbf{3.35E-02} &  7.86E-03 \\
        \textbf{Tai $et~al.$~\cite{tai2005local}} && 2.81E-02  & 2.77E+03  && 6.66E-02 &  2.12E-02\\
        \textbf{He $et~al.$~\cite{he2018deep}}    && 1.58E-01  & 4.88E+03  && 4.45E-02 &  9.13E-03 \\
        \textbf{Lee $et~al.$~\cite{lee2020deep}} && 3.65E-01  & 1.53E+03  && 4.51E-02 &  8.26E-03\\
        \midrule
        \midrule
        \textbf{Ours(R)}   && 1.44E-02  & 1.18E+03  && 5.17E-02 &  9.33E-03 \\ 
        \textbf{Ours}   && \textbf{4.49E-03}  & \textbf{1.04E+03}  && 3.53E-02 &  \textbf{7.54E-03} \\ 
        \bottomrule
    \end{tabular}
    }
\end{table}

\begin{table}[]
\centering
\caption{Image quality assessment with different metrics on images of VOC2012 reference by Zero-DCE images.}
\setlength{\tabcolsep}{3pt}    
\resizebox{\columnwidth}{!}{
\begin{tabular}{lcccccc}
\toprule 
Methods            & \textbf{Brisque} & \textbf{Clipiqa+} & \textbf{CNNIQA} & \textbf{DBCNN} & \textbf{MUSIQ} & \textbf{NIMA}  \\ \midrule
\textbf{Reinhard $et~al.$~\cite{reinhard2001color}} & 25.42            & 0.5764            & 0.7797          & 59.77          & 66.17 & 4.271          \\
\textbf{Tai $et~al.$~\cite{tai2005local}}      & 31.21   & 0.4413            & 0.8577          & 59.85          & 63.26          & 4.265          \\
\textbf{He $et~al.$~\cite{he2018deep}}       & 16.41            & \textbf{0.6014}   & -0.0556         & 60.01          & 63.66          & \textbf{4.782} \\
\textbf{Lee $et~al.$~\cite{lee2020deep}}      & 21.18            & 0.5867            & 0.7102          & 59.21          & 65.32          & 4.361          \\
\midrule
\midrule
\textbf{Ours(R)}     & 28.79        & 0.4446       & 0.9231 & \textbf{65.11} & 65.66          & 4.349   \\ 
\textbf{Ours} & \textbf{32.48} & 0.4861 &	\textbf{0.9711} &	64.75 &	\textbf{66.83} &	4.329 \\ 

\bottomrule        
\end{tabular}}
\label{T6_DCE}
\end{table}

\subsection{Quantitative Analysis}

\begin{figure}
  \centering
  \includegraphics[width=0.9\linewidth]{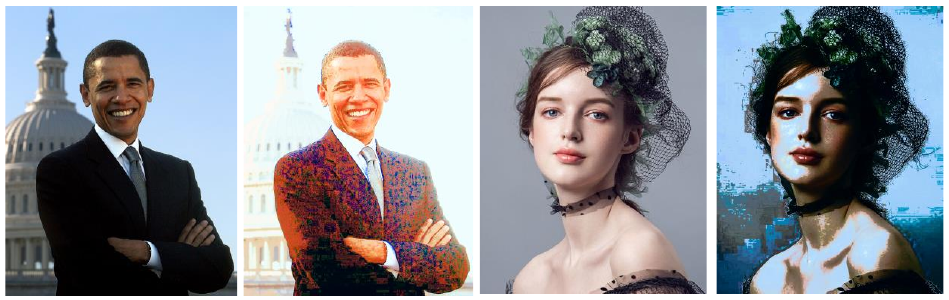}
  \caption{Some negative examples of color transfer judged by color consistency analysis. First example represents transfer result with bad $L$ channel-based consistency (abnormal exposure); second example represents transfer result with bad RGB-based consistency (color misalignment).}
  \label{fe4}
\end{figure}

The quantitative analysis for color transfer is still an open problem. In \cite{lee2020deep}, the author utilizes a subjective evaluation based on the user survey. Such evaluation takes too many uncontrollable factors in final result. We design some analysis tools to build stable quantitative analysis for color transfer. The tools include color consistency analysis and fading rate computation. The color consistency analysis is used to evaluate the degree of color deviation after color transfer. Some pixels 
with approximate RGB and $L$ values in source image should keep the similarity in transfer result. In Figure~\ref{fe4}, we show two negative instances (Obama and young lady portrait photos). Such similarities of the photos are broken by color transfer. To quantify the evaluation, we compute the histograms for RGB (10 bins for each main color) and $L$ (20 bins) channels from source image. Based on the histograms, the pixels can be classified into related bins. Then, we calculate the variance in each bin with corresponding RGB and $L$ values from transfer result. The average value of variances can be used to represent quantitative result of color consistency analysis. In RGB-based color consistency analysis, we achieve three average values from the three main color channels. For convenience, we average the three values again to obtain single value to represent the quality of RGB-based color consistency. In Table~\ref{T1_internet}, we show the quantitative analysis results of $L$ and RGB-based color consistency analysis based on image shown in Figures~\ref{fe1} and \ref{fe2}. It is clear that our method achieves better performance for color consistency keeping.

\begin{figure}
  \centering
  \includegraphics[width=0.9\linewidth]{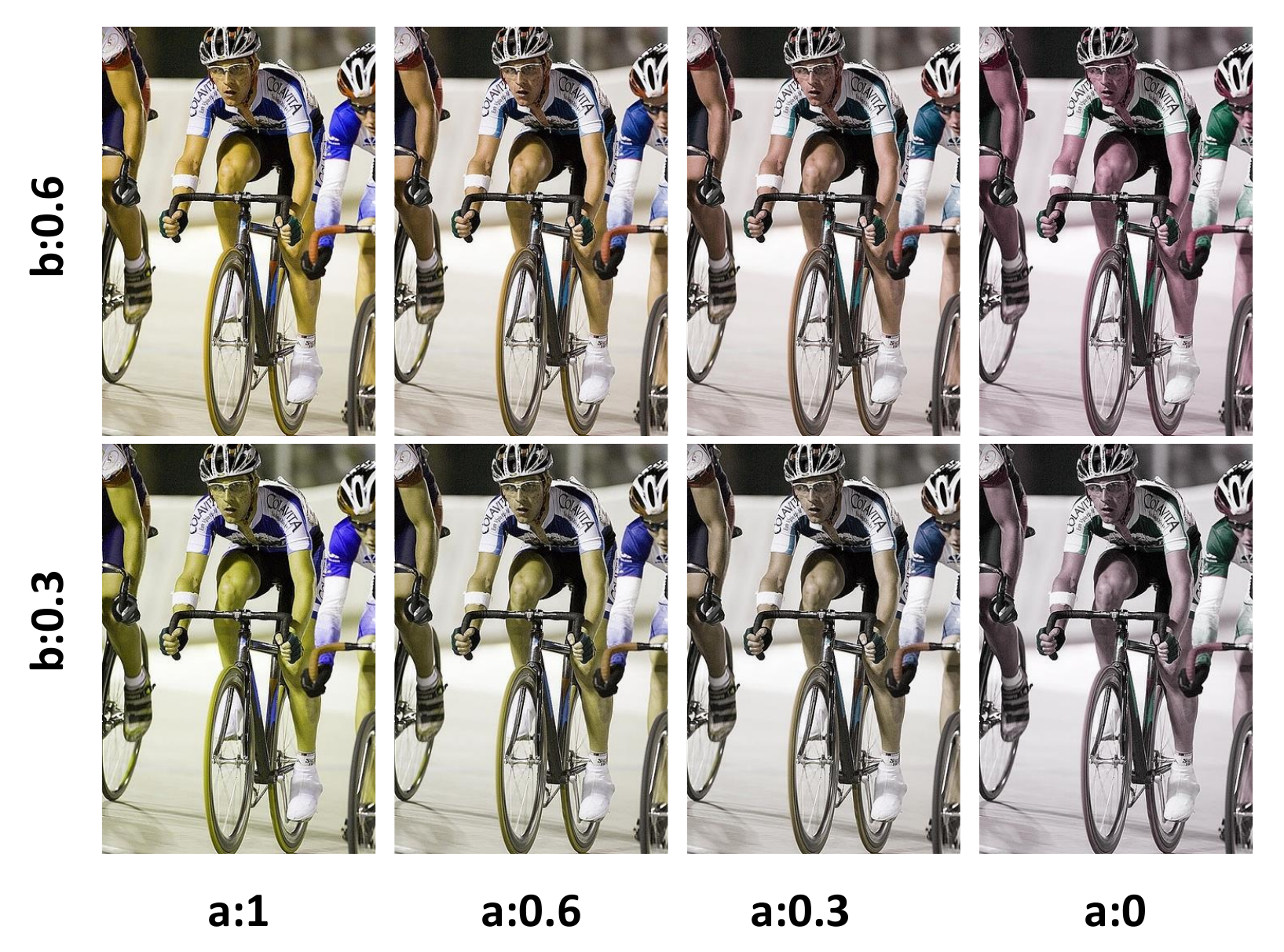}
  \caption{Instances of color degradation based on a-channel and b-channel. Horizontal axis: degradation (according to the product of the coefficient) in a-channel; vertical axis: degradation in b-channel.}
  \label{fe4.1}
\end{figure}

Another quantitative analysis tool is the fading rate computation. The main purpose of color transfer is to improve the quality of color distribution in source image. It can be regarded as a kind of image quality enhancement. If we consider the color distribution as signals, the strength of the signals should not be weakened by color transfer. A negative example is shown in Figure~\ref{fe4.1}. As the signals in a and b channels are weakened, the colorful image degrades into a gray-scale one. Based on the principle, we present the fading rate computation to quantitative represent the degree of color-based signal loss. We extract $a$ and $b$ values in Lab color space from source image and transfer result. The values are regarded as the color-based signals. We compute the difference of the values pixel by pixel between source image and transfer result to be the signal loss (when the value in transfer result is larger than source one, the signal loss is set to 0). Finally, we calculate the average from the signal loss to be the fading rate. In Table~\ref{T1_internet}, we report the results of fading rate for different methods based on images shown in Figures~\ref{fe1} and \ref{fe2}. Overall, our framework achieves better performance than the SOTA methods.

The proposed quantitative analysis tools are used to measure the quality of color transfer based on specific standards. In order to show the quantitative result for global quality evaluation, we introduce several classical image quality assessments, including Brisque~\cite{mittal2012no}, ClipIQA+~\cite{wang2022exploring}, CNNIQA~\cite{kang2014convolutional}, DBCNN~\cite{zhang2018blind}, MUSIQ~\cite{ke2021musiq}, and NIMA~\cite{talebi2018nima}. We implement the assessments on a subset with some inseparable landscape images. The global quality of color transfer can be better reflected on such images. In Table~\ref{T2_internet}, we show the quantitative results by the assessments based on images shown in Figures~\ref{fe1} and \ref{fe2}. Our method still achieve similar performance to the SOTA methods. 

\begin{figure}
  \centering
  \includegraphics[width=\linewidth]{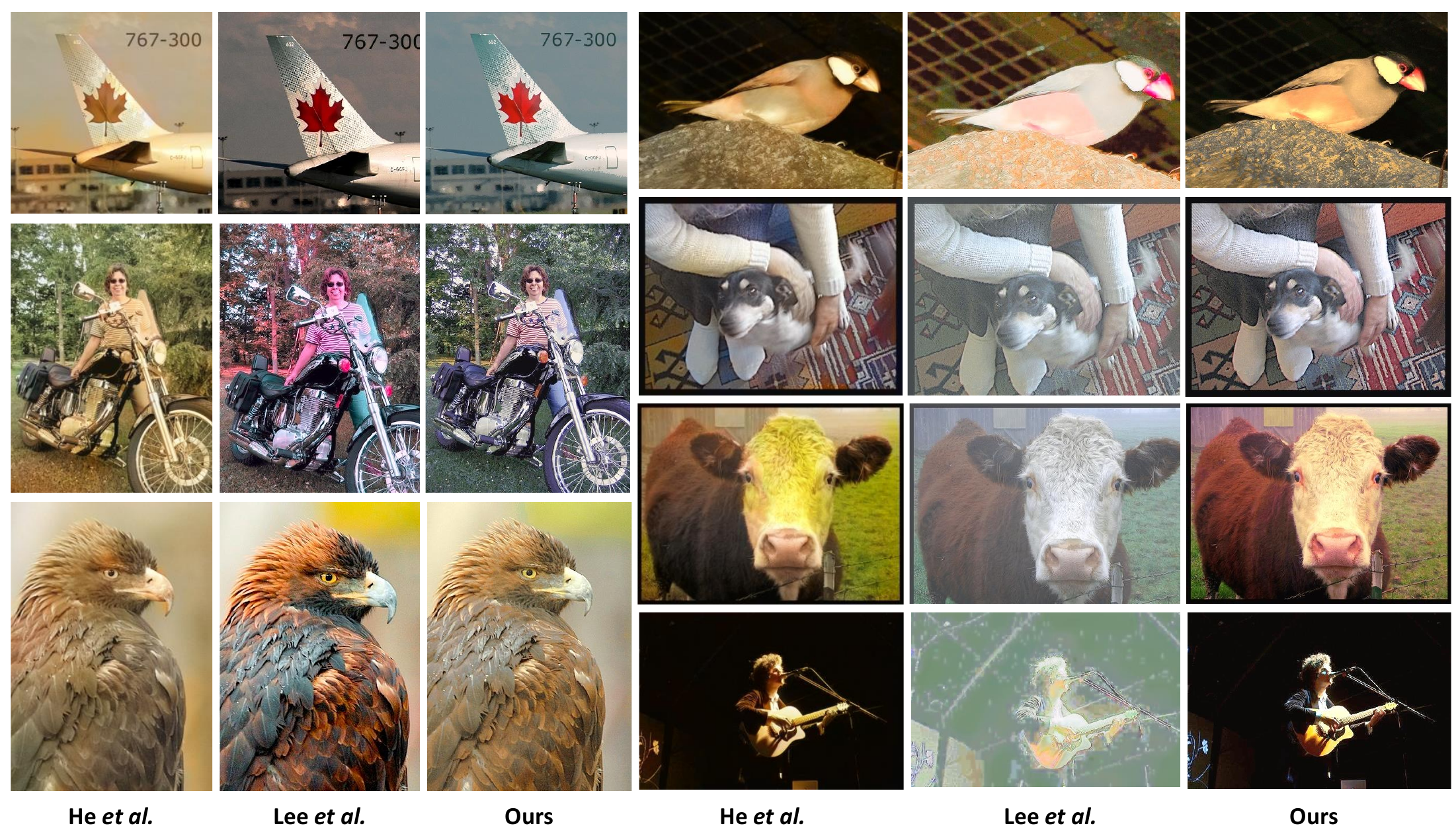}
  \caption{Comparisons between different color transfer methods on images of VOC2012. Methods: He $et~al.$~\cite{he2018deep} and Lee $et~al.$~\cite{lee2020deep}. With same reference images, our method achieves more natural color distributions.}
  \label{fe4_1}
\end{figure}

\begin{figure}
  \centering
  \includegraphics[width=\linewidth]{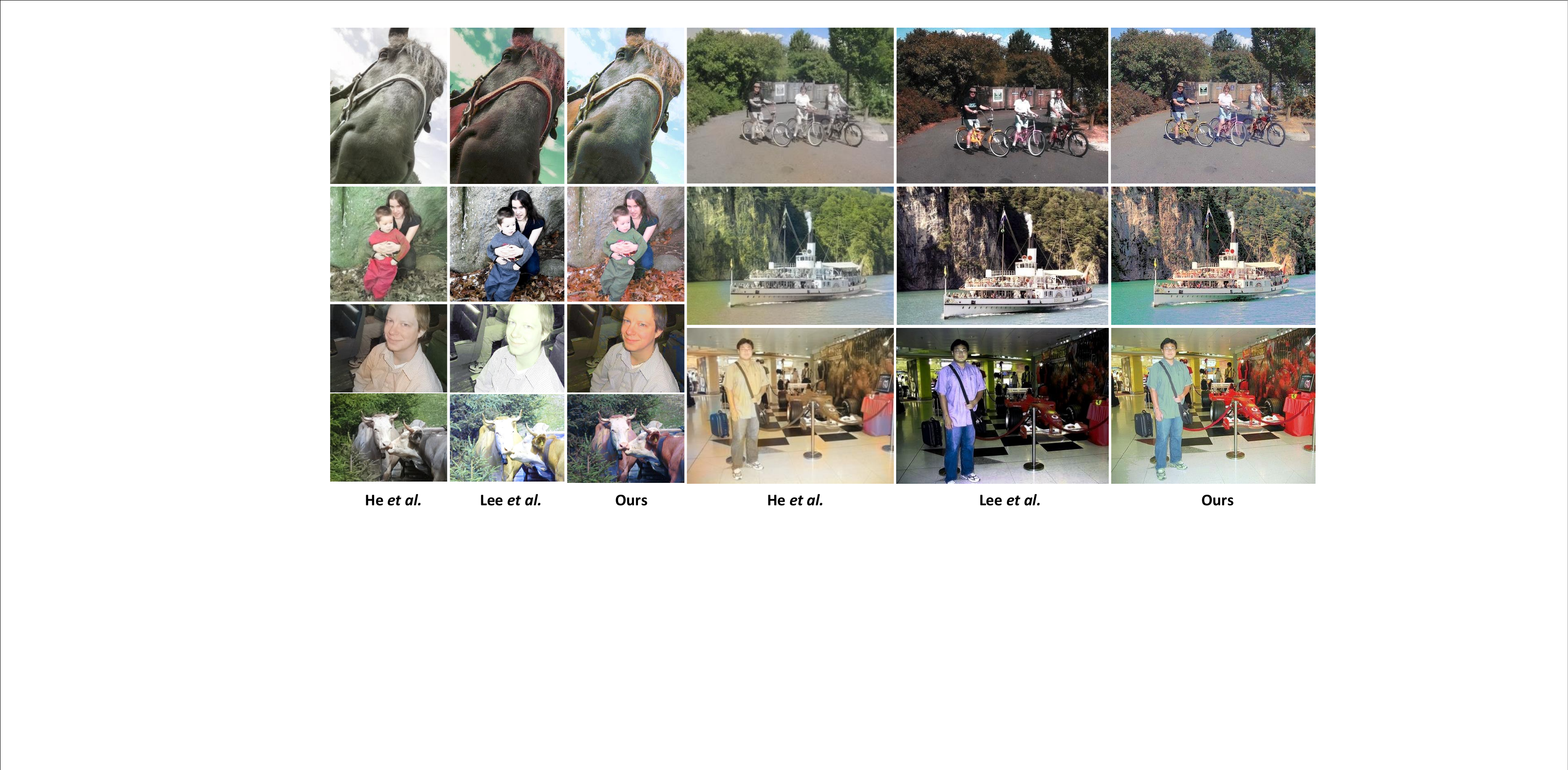}
  \caption{Comparisons between different color transfer methods on images of VOC2012 with reference images of Zero-DCE. Methods: He $et~al.$~\cite{he2018deep} and Lee $et~al.$~\cite{lee2020deep}. Our method achieves better exposure quality.}
  \label{fe4_2}
\end{figure}

The test images shown in Figures~\ref{fe1} and \ref{fe2} unavoidably take a certain degree of bias. To provide more persuasive evaluation, we collect two larger test datasets. One dataset contains 792 source images and 17 reference images from VOC2012 for separable picture testing. Related quantitative results are shown in Tables~\ref{T3_VOC} and \ref{T4_VOC}. Another dataset contains 1496 source images from VOC2012 and 8 reference images from Zero-DCE dataset~\cite{guo2020zero}. It is used to test the transfer performance in separable pictures (images of Zero-DCE dataset have not significant foreground and background). Same quantitative results are reported in Tables~\ref{T5_DCE} and \ref{T6_DCE}. Even the performance of our method is affected by complex correspondences and inseparable conditions in images, statistical advantages can still be demonstrated. As a simple ablation, we report transfer results of our method without lighting optimization (Ours(R) in all tables). It is clear that the lighting optimization improves the transfer quality, especially when the reference images take abnormal exposure regions. Some transfer results of SOTA methods and our framework are shown in Figures~\ref{fe4_1} and \ref{fe4_2}.

\begin{figure}[t]
  \centering
  \includegraphics[width=\linewidth]{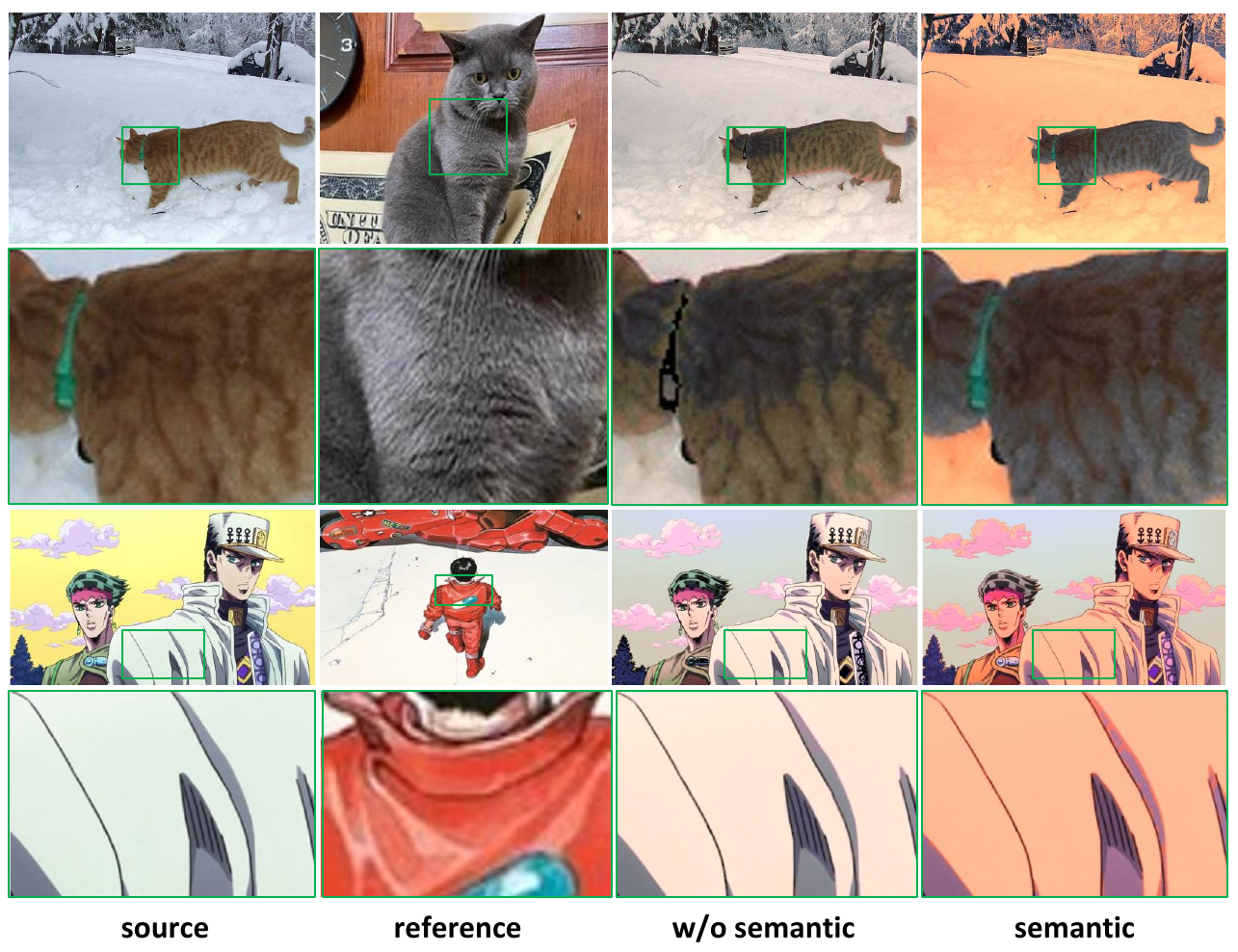}
  \caption{Comparisons of our color transfer without (w/o) and with semantic inspiration.}
  \label{fe4_3}
\end{figure}

\begin{figure}[t]
  \centering
  \includegraphics[width=\linewidth]{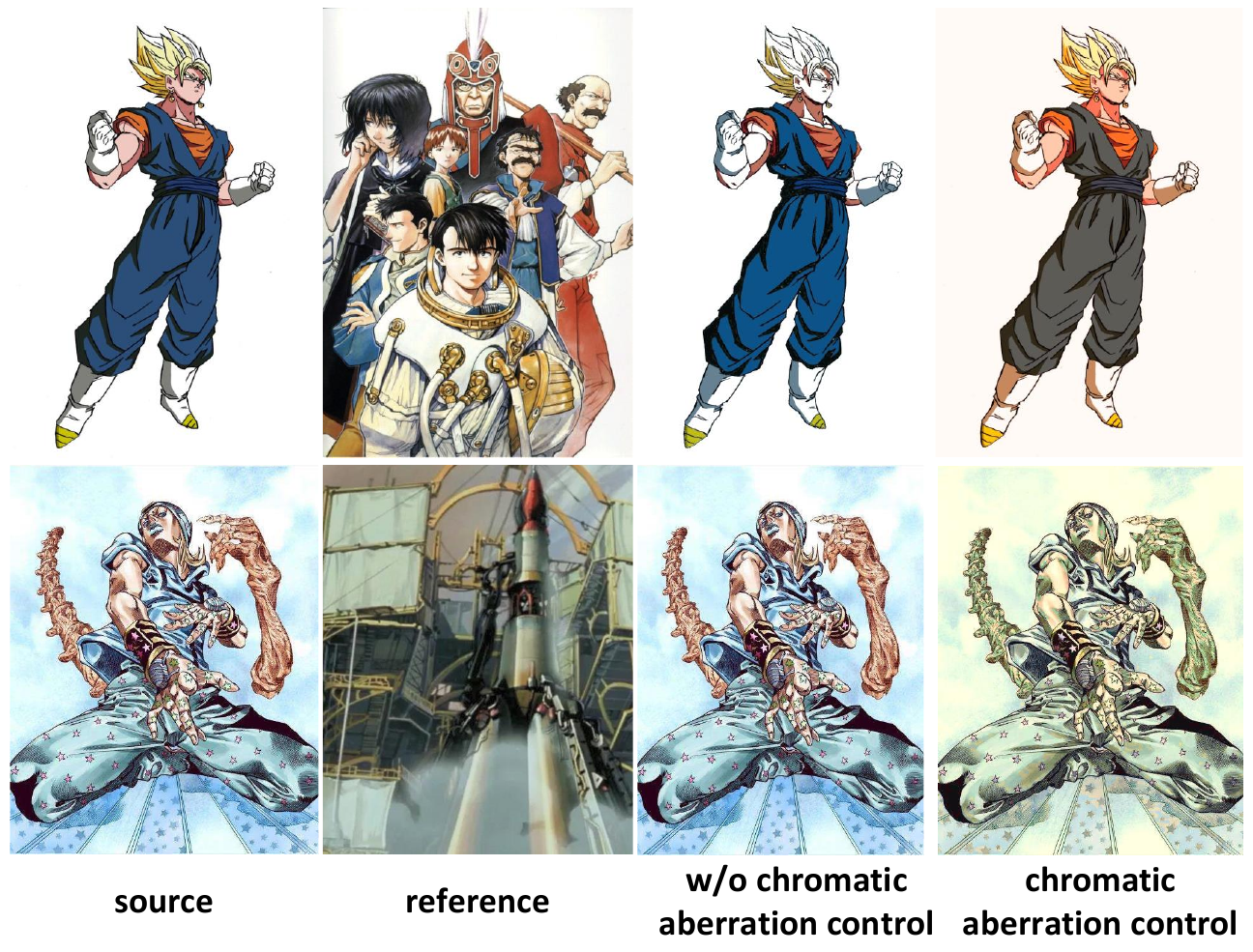}
  \caption{Comparisons of our color transfer without (w/o) and with chromatic aberration control.}
  \label{fe5_0}
\end{figure}

\subsection{Ablation}

The main steps of our method include palette-based clustering and color mapping, which have been meticulously designed and validated. The influence of the upper bound number and color mapping-based weighted update has been illustrated in Figures~\ref{f2} and \ref{f5}. To provide a more detailed explanation of the design, we elaborate on further details in the ablation study. Our method adopts a crucial idea: utilizing split correspondence. This idea serves as a semantic inspiration while avoiding specific object recognition, thereby reducing semantic sensitivity. To intuitively demonstrate this function, we compare several examples with and without the semantic inspiration in Figure~\ref{fe4_3}. It is evident that the semantic inspiration serves to align foregrounds and achieve more accurate color distributions. With the semantic inspiration, the semantic object of the foreground exhibits a more consistent transfer result according to the reference.

\begin{figure}[t]
  \centering
  \includegraphics[width=\linewidth]{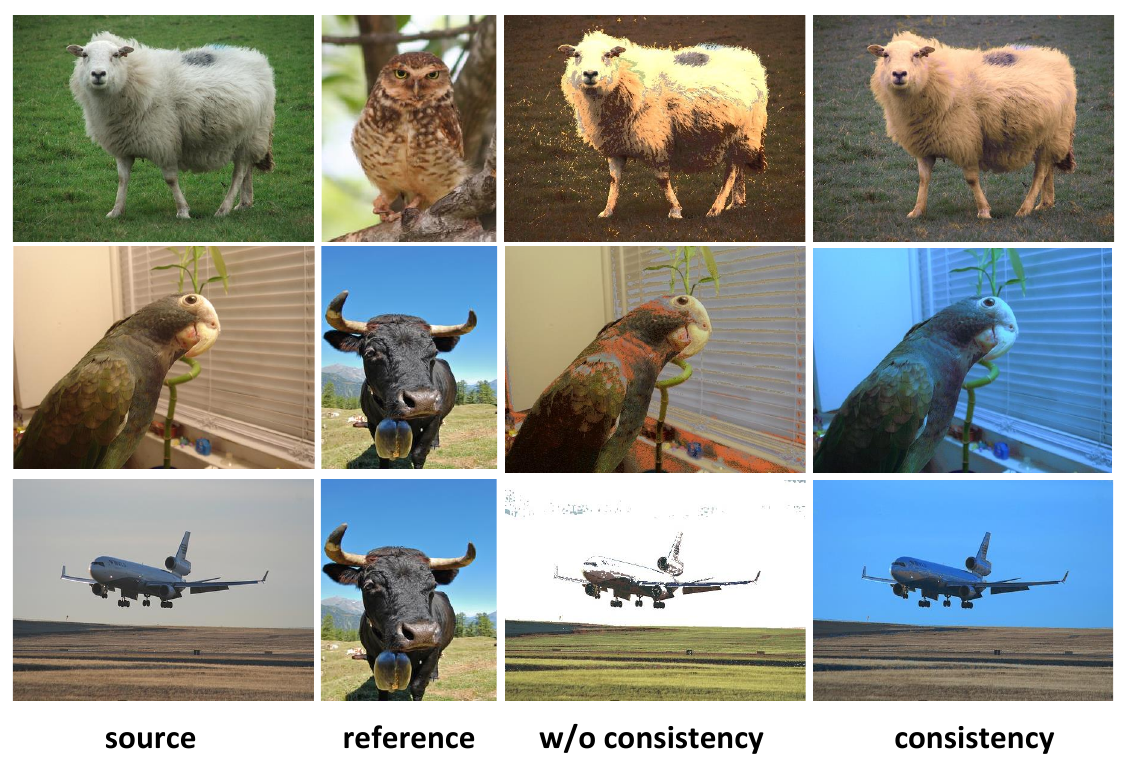}
  \caption{Comparisons of our color transfer without (w/o) and with color consistency.}
  \label{fe4_4}
\end{figure}

The maintenance of chromatic aberration control and color consistency are other critical aspects that require special attention. As we utilize the palette to establish key color-based correspondence with semantic inspiration, there arises a heightened demand for corresponding key colors and maintaining color consistency in the transferred images. In Figure~\ref{fe5_0}, we compare color transfer result without and with chromatic aberration control. It is clear that the chromatic aberration control is to align key colors between source and reference images. Without the control, the color distribution cannot be obviously changed according to the reference. As mentioned earlier, once the many-to-one key color mapping is established, it disrupts the structure of color distribution. Therefore, we maintain internal consistency and external continuity as described by Eq.\eqref{e7} and Eq.\eqref{e8}. Traditional palette-based strategies often change colors based on direct mappings of key colors without ensuring color consistency. This undoubtedly increases the likelihood of generating erroneous results. In Figure~\ref{fe4_4}, we demonstrate the color transfer outcomes using key color-based direct mapping alongside our method with color consistency. It is evident that without color consistency keeping, the transferred color distributions exhibit instability. Internal consistency and external continuity effectively prevent color overflow during the color transfer.

\begin{figure}[t]
  \centering
  \includegraphics[width=\linewidth]{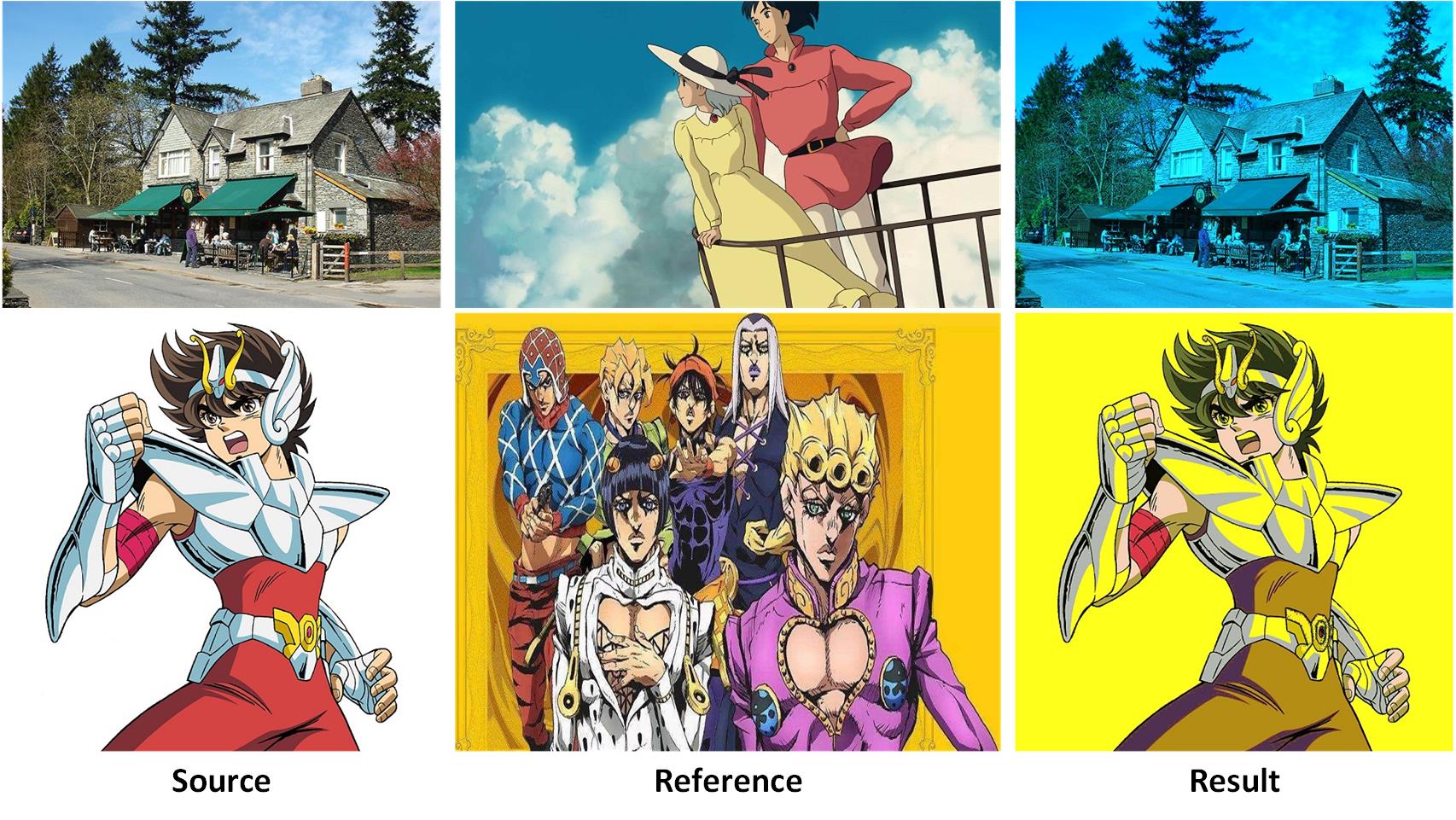}
  \caption{Some negative examples by our framework. The performance of the color transfer is degenerated into global mapping like filter effect.}
  \label{fe5}
\end{figure}

\color{black}
\subsection{Discussions}

The mentioned comparative results have proved that our framework has better performance in various color transfer tasks. For separable images, the proposed framework achieves more accurate color mapping for foreground and background parts. For inseparable images, our method still can implement stable result with global correspondence of color distribution while keeping the quality of color consistency. Comparing to the deep learning-based methods, our framework attempts to balance different factors and reduce the probability of incorrect color mapping, including color overflow, discontinuous color distribution, and abnormal exposure. 

Even our framework has many advantages for color transfer tasks, some limitations are still exist. Firstly, our method is sensitive to the palette-based correspondence between color distributions of images. If the color distributions are concentrated in a small range or the difference of the distributions are too large between source image and reference one, the transfer method may degenerate to a global color correspondence just like a color filter. In Figure~\ref{fe5}, two instances are shown. Another limitation is that our method cannot achieve real-time computing efficiency. The time cost of palette-based clustering and color transfer strategy is proportional to the scale of the images. In average, our method requires one to three minutes to transfer an image with $300\times300$ resolutions. The drawback restricts the application of our method for video processing.

\section{Conclusions}

In this paper, we propose a palette-based color transfer framework. It provides an applicable palette-based clustering to aggregate discrete color distributions by histogram analysis in Lab color space. Based on the palette, the proposed color mapping strategy utilizes split correspondence, chromatic aberration control, and color consistency keeping to achieve more accurate and robust transfer result. It guarantees the color consistency between reference image and transfer result while keeping the semantic consistency to the source image. With the lighting optimization, the production of abnormal exposure also can be controlled. The experimental results show that our framework achieves better transfer performance based on various evaluations. In future work, we will improve the efficiency of the method and introduce structural information analysis to inhibit transfer degradation.

\bibliographystyle{elsarticle-num} 
\bibliography{ref}





\end{document}